\documentclass[11pt,a4paper]{article}
\usepackage[hyperref]{naaclhlt2018}
\usepackage{times}
\usepackage{latexsym}

\usepackage{url}

\usepackage{enumitem}
\setlist{nolistsep}
\usepackage{amsmath}
\usepackage{amsfonts}

\usepackage{bm}

\usepackage{graphicx} 
\usepackage{subfig}

\aclfinalcopy 

\title{Contextual and Position-Aware Factorization Machines \\for Sentiment Classification}

\author{Shuai Wang \\
  University of Illinois at Chicago\\
  \hspace{-1mm}
  {\tt shuaiwanghk@gmail.com} \\\And
  Mianwei Zhou\\
  Yahoo! Research \\
  \hspace{5mm}
  {\tt mianwei@yahoo-inc.com} \\ \And
  Geli Fei\\
  University of Illinois at Chicago \\
  {\tt gfei2@uic.edu} \\ \AND
  Yi Chang\\
  Huawei Research America \\
  {\tt yichang@acm.org} \\ \\\And
  Bing Liu\\
  University of Illinois at Chicago \\
  {\tt liub@cs.uic.edu} \\
}

\begin{document}
\maketitle

\begin{abstract}

While existing machine learning models have achieved great success for sentiment classification, they typically do not explicitly capture sentiment-oriented word interaction, which can lead to poor results for fine-grained analysis at the snippet level (a phrase or sentence). Factorization Machine provides a possible approach to learning element-wise interaction for recommender systems, but they are not directly applicable to our task due to the inability to model contexts and word sequences. In this work, we develop two Position-aware Factorization Machines which consider word interaction, context and position information. Such information is jointly encoded in a set of sentiment-oriented word interaction vectors. Compared to traditional word embeddings, SWI vectors explicitly capture sentiment-oriented word interaction and simplify the parameter learning. Experimental results show that while they have comparable performance with state-of-the-art methods for document-level classification, they benefit the snippet/sentence-level sentiment analysis. 

\end{abstract}

\section{Introduction}
\label{sec:introduction} 
Although machine learning-based methods have achieved great success for sentiment classification~\cite{liu2012sentiment}, they have some limitations in explicitly capturing or presenting sentiment-oriented word interactions. Here the sentiment-oriented (SO) word interaction means that when two (or more) individual words function together as a sentiment expression (in a review snippet), the effect of word-wise interaction determines the sentiment orientation of that snippet. For example, in an online review snippet``this button is hard to push'', besides the sentiment polarity of every single word, the word interaction between ``hard'' and ``push'' indicates a negative signal for this snippet. Although some of the existing models consider such SO word interaction, it is coarsely modeled or implicitly captured, which we will detail in section~2. 
 
The lack of SO word interaction may not be critical for coarse sentiment classification at the document level (a full review), but the involvement of such SO interaction can play a crucial role in finer-grained analysis at the snippet level (a phrase or short sentence). The reasons are: 
(1) while a long document contains rich content, only limited text information is available in a short snippet, and (2) some salient opinion words (e.g., ``good'' and ``amazing'') can dominate document classification but these words may not always appear in short expressions classification.  

Specifically, the sentiment expression in a review snippet may consist of multiple words. Let us take a further look at the aforementioned example ``the button is hard to push''.  We can see the ``hard'' and ``push'' are used to deliver a negative opinion from a customer. However, ``\emph{hard}" or ``\emph{push}" independently indicate no clear sentiment. Notice that there are also other snippets like ``this is a hard (cellphone) case" where ``hard'' and ``case'' together assign a positive sentiment. In the above examples, an individual word like ``hard'', ``case'' or ``push'' is not able to determine the whole sentiment polarity of a snippet.
Instead, the word interactions play more important roles in identifying 
sentiment in such snippets, e.g., ``hard'' and ``push'' together indicate a negative opinion while ``hard'' and ``case'' interactively specify a positive opinion. 

This paper proposes a solution that can capture such interaction explicitly by exploiting Factorization Machine~\cite{rendle2010factorization}.
Factorization Machine (FM), 
which is 
widely used in recommender systems, is a general approach that can break the independence of interaction variables. It
suggests a possible way to realize our goal, i.e., to capture the SO interaction. However, direct application of FM is not suitable due to two main reasons. First, while FM aims at learning the global interaction between all elements, it neglects the importance of modeling (local) context in text. Different from recommender systems, contextual information plays a key part in sentiment analysis. Second, while the position/ordering of different features/fields in recommender systems may not be sensitive, the position information of words is an indicative signal in text data. 

To address them, we first propose Contextual Factorization Machine (CFM) which models context by capturing the focused interactions for a specific sentiment expression. After that we propose Position-aware Factorization Machine (PFM) to further encode position information. In these two models, the word interaction, context and position information are jointly learned by a set of vectors termed sentiment-oriented Word Interaction (SWI) vectors. Compared to word embeddings that are widely used in neural models for sentiment classification, SWI vectors explicitly capture SO word interaction and simplify the parameter learning. Experimental results show that while they give comparable performance with state-of-the-art methods for document-level classification, they effectively benefit the snippet-level analysis.

This paper makes the following contributions: 
\begin{enumerate}
\item It proposes a new solution to explicitly model sentiment-oriented word interaction for fine-grained sentiment analysis. 

\item It proposes two new models called CFM and PFM to learn a set of Sentiment-sensitive Word Interaction (SWI) vectors. Such vectors jointly capture word interaction, context and position information and also simplify the parameter learning.

\item Comprehensive experiments are conducted on three real-world review datasets at document and snippet/sentence level. By comparison with state-of-the-art models, experimental result shows the effectiveness of our approaches.\\
\end{enumerate}

\section{Sentiment-Oriented Word Interaction }
\label{sec:review}
In this section, we review some state-of-the-art machine learning methods for sentiment classification and analyze their limitations in modeling sentiment-oriented (SO) word interaction. Although some of them consider SO word interaction; however, it is coarsely modeled or implicitly captured. Based on the fashion of word representation, they are generally grouped into Bag-of-Words (BoW) and Word Embedding (WE) based methods. We illustrate them as follows and related notations are shown in Table~\ref{tab:notations}.


\subsection{Bag-of-Words (BoW) based Methods} 
In the Bag-of-Words model, words are indexed and text documents are converted to vectors. The values in such vectors can be word occurrence, word counts or TF-IDF. In this case, sentiment information is learned by corresponding model parameters. Specifically, in linear models like Logistic Regression (LR) in Equation~\ref{eq:logistic_regression} or non-linear models like SVM with feature projections (or kernels) in Equation~\ref{eq:svm}, we can see $\bm{w}$ is the parameter capturing sentiment information under supervised learning. For simplicity of illustration, bias terms are excluded here but they will still be used in our experiments.


\begingroup
\vspace{-3mm}
\begin{equation} 
\label{eq:logistic_regression}
 LR: y = \sigma (\bm{w} \cdot \bm{x}) 
\end{equation}
\vspace{-5mm}
\begin{equation} \label{eq:svm}
	SVM: y = \langle \bm{w},\phi (\bm{x})\rangle 
\end{equation}
\endgroup

When BoW is used in linear models like LR, the sentiment information captured by $\bm{w}$ is interpretable but we cannot measure the direct interaction between words. For example, a learned parameter of the word ``good" (i.e., $w_{good}$) can make a text snippet containing ``good" more likely to be predicted as positive (when $w_{good}$ $>$ 0), which is straightforward. However, it has problems when predicting sentiment for snippets containing ``hard'' and ``push'' or ``hard'' and ``case''. Ideally, their word-wise interaction should be considered but here the sentiment polarity is determined by $ w_{hard},w_{push}, w_{case}$ which are independent variables. On the other hand, when BoW is used with non-linear projection like SVM with non-linear kernels, it may help solve the problem but it is harder to track the SO word interaction due to the non-linear feature projections. 


\noindent
\textbf{SVM with Polynomial Projection (SVM-Poly)}\\
Using BoW with a $m$-degree polynomial feature projection (or $m$-poly kernel) is an exception, for example, the 2-poly kernel. Its feature mapping and prediction function are shown in Equation~\ref{eg:poly2} and~\ref{eg:poly2_enpanded}. Note that the bias and linear terms are excluded for simplicity in Equation~\ref{eg:poly2_enpanded}. One can see this approach is capable of capturing word interactions, for instance, the direct interaction of ``hard'' and ``case'' can be parameterized as $w_{hard, case}$. However, this is still problematic as all such interaction parameters are independent. That is, we can learn $w_{hard, case}$ and $w_{hard, push}$ but they are two isolated parameters, regardless of the fact that they share the word ``hard''. In addition, this approach suffers from data sparsity and we need $N^{m}$ parameters and $O(N^{m})$ time complexity, specially when the key words for interaction are distant, like ``hard'' and ``case'' in ``the case I bought is really hard''. Finally, it is worth noting that the problems of $m$-gram BoW model ($m > 1$) are very similar to the ones of SVM-Poly by shifting the encoding of word-pairs from $w_{hard, case}$ to $x_{hard, case}$. For consistency, we will use SVM-Poly as a general case for the following discussion.


\begingroup
\begin{equation} 
\label{eg:poly2}
\begin{split}
\phi_{poly2} (\bm{x}) & = (1,\sqrt 2 {x_1},...\sqrt 2 {x_n},{x_1}^2,...,{x_1}^n, \cr
&\sqrt 2 {x_1}{x_2},...,\sqrt 2 {x_2}{x_3},...,\sqrt 2 {x_{n - 1}}{x_n})\cr
 y &  = \langle w,\phi_{poly2} (\bm{x})\rangle \cr
\end{split} 
\end{equation}
\begin{equation} 
\label{eg:poly2_enpanded}
y =  \sum\nolimits_{i = 1}^n {\sum\nolimits_{j = i + 1}^n {{w_{i,j}}} {x_i}{x_j}}
\end{equation} 
\endgroup

\subsection{Word Embedding (WE) based Methods} 
Recently word embeddings become widely used in many machine learning models. Generally speaking, the embeddings for words are mainly learned by maximizing the likelihood of correct prediction of contextual information, e.g., the skip-gram model~\cite{mikolov2013efficient}.
Consequently, words that are semantically similar have similar representations, e.g., ``cost'' and ``price''. 
However, such word embeddings do not directly carry sentiment information, e.g., ``good'' and ``bad'' are also neighbors\footnote{Observed from the default result generated by word2vec.} in word vector space but they actually hold opposite sentiment polarities.

\renewcommand{\arraystretch}{0.9} 

\begin{table}
\small
\begin{tabular}{|cl|}
\hline
$x$ & a word feature, e.g., word occurrence \\
$\bm{x}$ & a sequence of word features \\
$\phi(x)$ & feature mapping/projection \\
$y$ & a prediction, e.g., sentiment polarity\\
$\sigma $ & sigmoid function\\
$v$ & a word vector/factor \\
$w$  & a model parameter for learning  \\
$b$ & bias term \\
$f $ & a non-linear activation function\\
$ k $ & dimension of a factor/vector \\
$ t $ &  distance (context size)   \\
$ l $ & number of convolutional layers \\
$ g $ & number of channels \\
$ h $ &  word window size (length of a CNN filter) \\ 
$ n $ & average number of words in a document\\ 
$ N_d$ &  number of words in document d \\
$ N $ & number of words in vocabulary\\
$ ds(i,j) $ & distance between two words i and j\\
$ g(i,k) $ & gradient of v(i,k)\\
$\lambda$&  regularization term\\
 $\eta$  & learning rate \\
     \hline
\end{tabular}
\caption{\label{tab:notations} Definition of Notations}
\end{table}

A natural way to learning sentiment information with WE is following the manner of BoW, i.e., to train a classifier like LR/SVM.
Certainly, deep learning models like convolutional neural network (CNN) can be more suitable to employ word embeddings as input with their particular architecture designs. Recently some advanced models jointly encode semantic and sentiment information in word vector space~\cite{kim2014convolutional, tang2014learning}. However, 
they still do not explicitly reflect SO word interaction. 

The reason is, while most of the neural network models are based on full sentence/document modeling, they are coarse-grained in nature and not good at capturing fine-grained information at the word/snippet level~\cite{he2016pairwise}. Specifically, let us first investigate the convolution operation in Equation~\ref{eg:dp}. Note that a word is now denoted by a vector $v_{i} \in {\mathbb{R}^k}$. The learning parameter $w$ is applied to a window of $h$ words to generate a convolutional feature $c_{i}$. A feature map $\bm{c}$ is then obtained by processing a sequence of words. After that, a max pooling operation is applied to take the maximum value~\cite{collobert2011natural} for each such feature map. 

Although here $\bm{w}$ can capture the interaction between different words, it is not used in a word-wise manner, i.e., the SO word interaction is modeled implicitly, as is is hard to measure the direct interaction between two particular words, say ``hard'' and ``case''. However, they can be explicitly encoded in $w_{hard, case}$ in SVM-Poly. Also, due to the non-linear function $f$ and pooling operation in CNN, the specific word interaction between two words becomes harder to track.
 
\begingroup
\begin{equation} 
\label{eg:dp}
\begin{split}
& {c_i} = f(\bm{w} \cdot {\bm{v_{i:i + h - 1}}} + b) \cr
& \bm{c} = [{c_1},{c_2},...,{c_{n - h + 1}}] \cr
\end{split} 
\end{equation}

\endgroup

\section{Proposed Factorization Machines}
\label{sec:models}
As discussed above, some related models coarsely consider SO word interaction and they have limitations. For example, CNN considers the contextual relationship between words and encodes all interactions in a general parameter $w$ but the specific word-wise interaction is implicitly modeled and hard to track. SVM-Poly encodes SO word interaction in parameters like $w_{i,j}$ but it requires $O(N^{m})$ such parameters and those parameters are all independent. 

It will be a promising direction if we can adopt their advantages while overcome their shortcomings in a joint modeling process. Motivated by this, Factorization Machine (FM) is exploited by us. However, notice that FM is originally used in recommender system and does not directly applicable for fine-grained sentiment analysis. Therefore, we propose two new models CFM and PFM.

In this section, we first introduce the basis of FM. We then illustrate how to exploit it and point out its problems in fine-grained sentiment analysis. After that, we represent our new models, the optimization approach, and the  analysis of complexity.


\subsection{Factorization Machine Basis}
\label{label:fm}
Factorization Machine (FM) ~\cite{rendle2010factorization} was proposed as a generic framework to learn the dependency of interaction variables by factorizing them into latent factors. A factor can be generally understood as a vector. So in this paper we will use the term \emph{factor} and \emph{vector} interchangeably. The model equation of 2-degree factorization machine is presented in Equation~\ref{eg:fm}. Here $v_{i}$ is called a factor/vector for element $x_{i}$ ($v_{i} \in {\mathbb{R}^k}$) and $\langle v_{i}, v_{j} \rangle$ denotes the dot product operation of $v_{i}$ and $v_{j}$. 
Similarly, its linear and bias terms are not included here but will be used in experiments. 

\begingroup
\begin{equation} 
\label{eg:fm}
\begin{split}
& y = \sum\nolimits_{i = 1}^n {\sum\nolimits_{j = i + 1}^n {\langle {v_i},{v_j}\rangle } {x_i}{x_j}} \cr
\end{split} 
\end{equation}
\endgroup

\subsection{Exploiting FM for Sentiment Analysis}
This is for test. 
We exploit FM for sentiment analysis in the following manner: while $x_i$ is used as word features like LR/SVM for the word at position $i$, the factor $v_{i}$ can be viewed as its word vector. However, different from the traditional word embedding, the word vector here carries word-wise interaction information and is sentiment-sensitive. We refer it as sentiment-sensitive Word Interaction (SWI) vector. In this setting, the SO interaction between two words is determined by the dot product of their SWI vectors, for example, the interaction between ``hard'' and ``case'' is denoted by $\langle v_{hard}, v_{case} \rangle$. 

Let us compare FM (Equation~\ref{eg:fm}) and SVM-Poly (Equation~\ref{eg:poly2_enpanded}) for a better understanding. By comparison, one can see that instead of using an independent interaction parameter $w_{i,j}$, here the SO interaction effect of two words is jointly determined by two SWI vectors $ {v_i}$ and ${v_j}$. Recall that $w_{hard, case}$ and $w_{hard, push}$ are two isolated parameters in SVM-Poly, but in our case, $\langle v_{hard}, v_{case} \rangle$ and $\langle v_{hard}, v_{push} \rangle$ can reflect that they share the same word ``hard'', because they both contain the SWI vector~$ v_{hard}$.  Meanwhile, note that their resulting sentiment polarities are different, when $v_{hard}$ interacting with $v_{push}$ is probe to the negative class (e.g., $\langle v_{hard}, v_{push} \rangle \approx 0$, where $0$ indicate a negative sentiment class) and $v_{hard}$ interacting with $v_{case}$ is close to positive class (e.g., $\langle v_{hard}, v_{case} \rangle \approx 1$). 
The SWI vectors such as $v_{hard}, v_{case}$ and $v_{push}$ are jointly learned under the supervision of sentiment labels.

However, this direct application of FMs is not suitable for fine-grained analysis at the snippet level due to two main reasons: (1) they lack the modeling of contextual information; (2) they do not consider the position/ordering of words. But they are two important signals to connect aspect and opinion information for sentiment reasoning at the snippet level. To address them, two new models are proposed and introduced below.

\subsection{Contextual Factorization Machine}
\label{sec:cfm}
We first propose Contextual Factorization Machine (CFM). Different from other existing FMs, CFM models contextual information in text. Note that in fine-grained analysis, we aim at detecting sentiment for an aspect-specific opinion expression, for example, a snippet \textbf{\emph{the screen is very clear}} from a full review ``I made this purchase two days ago ... the screen is very clear ... ''. We observe that to generally determine the sentiment of that snippet, there is no need to catch fully pairwise word interactions. Concretely, the sentiment interaction between word ``screen'' to the first/last few words in the original full review could be less informative. Those words may even not be related to ``screen'' but another aspect (e.g., ``purchase''). As a result, their word interactions can be harmful if they are involved in learning. So an intuitive solution is to focus on the interactions constructed by nearest contextual words. In other words, by capturing the most significant word dependency, CFM can learn fine-grained SO interaction more accurately. Its model equation is shown in Equation~\ref{eg:cfm}. The idea is to impose a constraint $t$ so that word interactions will be learned within a certain distance, which is inspired by the neural skip-gram model. However, here it is designed for better alignment of aspect and opinion information but not for word prediction.

\begingroup
\begin{equation} 
\label{eg:cfm}
\begin{split}
y  & = \sum\nolimits_{i = 1}^n {\sum\nolimits_{j = i + 1}^{\min (i + t,{N_d})} { \langle {v_i},{v_j} \rangle {x_i}{x_j}} } \cr
& = {1 \over 2}\sum\nolimits_{l = 1}^k \Big( {\sum\nolimits_{i = 1}^n {\sum\nolimits_{j = \max (1,i - t)}^{\min \left( {i + t,{N_d}} \right)} { {v_{i,l}},{v_{j,l}} {x_i}{x_j}} } }\cr
& { - \sum\nolimits_{i = 1}^n { {v_{i,l}},{v_{i,l}} } {x_i}{x_j}} \Big) \cr
\end{split} 
\end{equation}
\endgroup

\subsection{Position-aware Factorization Machine}
\label{sec:dfm}
One shortcoming of CFM is that it considers the same words with different word positions identical in SO interaction, which is not always true. In fact, word positions may be helpful to distinguish different sentiment polarities in some cases. To incorporate this indicative signal, we create a more comprehensive model named Position-aware Factorization Machine (PFM), where the SO word interaction, context and position information are jointly learned by the SWI vectors. Equation~\ref{eg:dfm} shows its model equation. Compared to CFM, $ds(i,j)$ is newly-designed to denote the distance between two words.
Take the snippet ``the case is very hard'' again as an example and we 
will have $ds(case, hard)=3$, i.e., the distance between ``case'' and ``hard'' is 3, and now their SO word interaction is depicted as $\langle v_{case,3}, v_{hard,3} \rangle$. 

\begin{multline}
\label{eg:dfm}
y  = \sum\nolimits_{i = 1}^n {\sum\nolimits_{j = i + 1}^{min(i + t,{N_d})} {\langle {v_{i,ds(i,j)}},{v_{j,ds(i,j)}}} } \rangle {x_i}{x_j}\\
= {1 \over 2} \sum\nolimits_{l = 1}^k \Big(  \sum\nolimits_{i = 1}^n {\sum\nolimits_{j = \max (1,i - t)}^{\min (i + t,{N_d})}}\\ 
{  {v_{i,ds(i,j),l}},{v_{j,ds(i,j),l}} {x_i}{x_j}} \\
{ - \sum\nolimits_{i = 1}^n { {v_{i,ds(i,j),l}},{v_{i,ds(i,j),l}}  {x_i}{x_i}} } \Big) 
\end{multline}



\subsection{Optimization}
In terms of learning, we formulate our task as an optimization problem. Since the sentiment information needs to be learned by supervision from document labels (positive/negative), we use logistic loss to optimize. In addition, we impose L2 regularization parameterized by $\lambda$. Following~\cite{jahrer2012ensemble}, mini-batch based stochastic gradient descent (SGD) is used. We also implement the adaptive learning-rate schedule~\cite{zeiler2012adadelta} to boost our learning process. Particularly, AdaGrad~\cite{duchi2011adaptive} is adopted. We show the gradient of the factor $v_{i}$ in CFM below and the gradient for PFM can be derived similarly. $\lambda$ is the regularization term and $\eta$ is the learning rate.

\begingroup
\begin{equation} 
\label{eg:pfm}
\centering
\begin{split}
  {g_{i,l}} & \leftarrow  \sum\nolimits_{j = \max (1,i - t),j \ne i}^{\min (i + t,{N_d})} {{v_{j,l}}{x_i}{x_j}}  + \lambda {v_{i,l}}\cr
  {G_{i,l}}& \leftarrow {G_{i,l}} + {g_{i,l}}^2 \cr
  {v_{i,l}} & \leftarrow {v_{i,l}} - {\eta  \over {\sqrt {{G_{i,l}}} }}{g_{i,l}}\cr
\end{split} 
\end{equation}
\endgroup

\subsection{Complexity and Analysis}
We report the number of parameters and complexity for learning in Table~\ref{tab:complexity}. $n$ is the average length of one document. $t$ is the distance indicator and $k$ is the vector dimension. CNN-S means the CNN model using static word embeddings as input and CNN-J means the CNN model jointly learning word embeddings. The meaning of other symbols can be found in~Table~\ref{tab:notations}. We have the following observations: (1) both CFM and PFM are linear in $n$ for the growth of variables and complexity. (2) CFM and PFM are faster than FM because FM calculates all pairwise interactions\footnote{The proof of $O(nk)$ for FM is reported in\cite{rendle2010factorization}} while they do not need to. (3) CFM and PFM are both less complicated than SVM-Poly. While $O(N^{2})$ parameters are required to learn all pairwise word interactions in SVM-Poly, only $O(Nk)$ ones are needed for FMs. (4) Compared to CNNs, CFM and PFM simplify the learning process. That is because while word embeddings are used in the input layer and CNN learns the sentiment information by other deep layers, the sWI vectors used in CFM and PFM jointly encode all related information. 


\renewcommand{\arraystretch}{0.9} 
\begin{table}
\small
\begin{center}
\begin{tabular}{|c|c|c|c|c|c|c|c|c|}

\hline
\rule{0pt}{8pt} \textbf{Model} &\textbf{Parameters} & \textbf{Time Complexity}
\\
\hline

LR/SVM & $n$ & $O(n)$\\ \hline
FM& nk & $O(nk)$ \\ \hline
CFM& nk  & $O(nk)$ \\ \hline
PFM & nkt  & $O(nk)$ \\ \hline
SVM-Poly & $n^2$ &$O(n^2) $ \\ \hline
CNN-S & $nk+lhkg$ & $O(lhkg) $\\ \hline
CNN-J & $nk+lhkg$ & $O(nk+lhkg)$ \\ \hline
\end{tabular}
\end{center}
\caption{\label{tab:complexity} Comparison of the number of parameters and computing complexity with related models}
\end{table} 

\section{Experiments}
\label{sec:experiment}
Our evaluation is two-step. First, we conducted sentiment classification at the document level using full online reviews. Second, we used the models built from full reviews to classify review snippets (phrases or sentences). Specifically, a set of review snippets with human labels (positive/negative) was used as our prediction targets while we still utilized the same set of full documents for training. The intuition is that, as discussed in section~\ref{sec:introduction}, the SO word interaction may have limited impact at the document level, but it plays a crucial role for fine-grained analysis at the snippet level, because a short text usually contains limited information or has less strong salient opinion words (e.g., ``excellent''). 
In this case, when all candidate models are trained on a same set of full documents, a model better capturing explicit SO word interaction should be able to identify the sentiment of a short snippet more accurately.



\subsection{Datasets}
\label{subsec:datasets}
We use three real-word review datasets. The label for a full review can be directly obtained because a rating score is often provided, but the label for a text snippet requires human labeling. We thus download the human-labeled snippets from the UCI dataset\footnote{https://archive.ics.uci.edu/ml/machine-learning-databases/00331/} as our test sets in our second step. They are word phrases or short sentences (snippets from full reviews) about movie from IMDB, cellphone from Amazon and restaurant from Yelp. Each set contains 1000 snippets (500 positive/negative).
For full reviews, we use the movie review dataset from~\citet{pang2004sentimental} which contains 1000 positive and 1000 negative movie reviews, cellphone reviews from~\citet{mcauley2015image}, and restaurant reviews from Yelp~\footnote{http://ww.yelp.com/dataset\_challenge}. For consistency, 1000 positive and 1000 negative reviews are sampled from cellphone and restaurant. Reviews with rating scores 5 and 4 are treated as positive and scores 2 and 1 are treated as negative like~\cite{chen2015lifelong, johnson2014effective}. For the first task, the document-level sentiment classification, we conduct 5-folds cross validation using only full reviews. We split the data to 70\%, 10\%, 20\% for training, validation, and testing for each data set. For the second task, we use the classifiers trained by all full reviews to classify review snippets. Notice that we have kept the same parameter settings for the classifiers used in both tasks. The average document/snippet length of each data set is reported in Table~\ref{tab:dataset}.

%
%
\renewcommand{\tabcolsep}{1.5pt} 

\begin{table}
\small
\begin{center}
\begin{tabular}{|c|c|c|c|c|c|c|c|c|}

\hline
\multicolumn{2}{|c|}{\rule{0pt}{8pt} \textbf{Data}} & \multicolumn{1}{|c|}{\rule{0pt}{8pt} \textbf{Training (docs)}} & \multicolumn{1}{|c|}{\rule{0pt}{8pt} \textbf{Testing (snippets)}} \\
\hline
\rule{0pt}{8pt} \textbf{Dataset} &\textbf{Source} & \textbf{Average Length}  & \textbf{Average Length}
\\
\hline

Cellphone  & Amazon  & 90 & 10\\
\hline
Restaurant &  Yelp &  70 & 11  \\
\hline
Movie & IMDB    & 668 & 15  \\
\hline
\end{tabular}
\end{center}
\vspace{-2mm}
\caption{\label{tab:dataset} Data Information}
\vspace{-5mm} 
\end{table} 

\subsection{Candidate Models for Comparison}
\textbf{FM}: The classic factorization machine.\\
\textbf{CFM}: Contextual Factorization Machine.\\
\textbf{PFM}: Position-aware Factorization Machine. \\
\textbf{SVM-BoW}: Linear SVM with Bag-of-Words.\\
\textbf{SVM-Poly}: SVM with the Poly kernel. \\
\textbf{LR-BoW}: Logistic regression with Bag-of-Words, similar to SVM-BoW in settings.\\
\textbf{SVM-WE}: SVM with word embeddings. The averaged feature values of word embeddings are used for review representations. This is a setup to evaluate the embedding contributions ~\cite{ding2015mining, kotzias2015group}. We use word2vec\footnote{https://code.google.com/archive/p/word2vec/} to train vectors for every dataset. \\
\textbf{LR-WE}: Logistic regression with word embeddings, similar to SVM-WE in settings.\\
\textbf{CNN-S}: Convolutional Neural Network, a representative neural model for sentiment classification using word embeddings~\cite{kalchbrenner2014convolutional,kim2014convolutional}. The word embeddings are pre-trained and used as static input.\\
\textbf{CNN-J}: another CNN whose word embeddings will be jointly learned with other parameters in the deep layers~\cite{kim2014convolutional,tang2014learning}. We put pre-trained word embeddings for initialization.\\
\textbf{CNN-S(+)}: Similar to CNN-S, but we increase the length of word embedding to 300  (i.e., longer vectors), while the length of word embedding in above CNNs is maintained the same as in all FMs.\\
\textbf{CNN-J(+)}: Similar to CNN-J, but we increase the word embeddings length following CNN-S(+).\\

\subsection{Parameter Settings}
For CFM and PFM, we set the context size $t$ to 5 following related vector learning approaches~\cite{mikolov2013distributed, mnih2013learning} and the dimension of word vector $k$ to 10. We did pilot experiments and found a bigger vector length does not have significant influence, which indicates that to learn SWI vectors a small vector length is enough. We maintain the same setting for training skip-gram vectors which are used in SVM-WE, LR-WE, CNN-S and CNN-J. We learn vectors with a large length $k = 300$~\cite{mikolov2013distributed} for CNN-S(+) and CNN-J(+). The learning rate $\eta$ is empirically set to 0.01 for CFM and PFM. 
The regularization term $\lambda$ is set to 1 for CFM for all data sets, which works consistently well. 
Bias and L2 regularization terms are also used in SVM-BoW, LR-BoW, SVM-WE and LR-WE for consistency. We follow the parameter settings from~\citet{kim2014convolutional} for CNNs.

%
%
\renewcommand{\tabcolsep}{1.5pt} 

\begin{table}
\small
\begin{center}
\begin{tabular}{|c|c|c|c|c|c|c|c|c|}

\hline
\multicolumn{1}{|c|}{\rule{0pt}{8pt} \textbf{}} & \multicolumn{2}{|c|}{\rule{0pt}{8pt} \textbf{Document}} & \multicolumn{2}{|c|}{\rule{0pt}{8pt} \textbf{Snippet}} \\
\hline
\rule{0pt}{8pt} \textbf{Models}  & \textbf{Accuracy}  & \textbf{F1-Score}  & \textbf{Accuracy}  & \textbf{F1-Score}
\\
\hline

LR-BoW	&	\textbf{0.863}	&	\textbf{0.862}	&	0.711 	&	0.709 	\\	
SVM-BoW	&	0.855 	&	0.855 	&	0.709 	&	0.706 	\\	
SVM-Poly & 0.832 & 0.832 & 0.500 & 0.333 \\ 
LR-WE	&	0.661 	&	0.661 	&	0.654 	&	0.649 	\\	
SVM-WE	&	0.667 	&	0.666 	&	0.658 	&	0.652 	\\	
CNN-S	&	0.683 	&	0.676 	&	0.601 	&	0.555 	\\	
CNN-J	&	0.747 	&	0.745 	&	0.609 	&	0.570 	\\	
CNN-S(+)	&	0.851 	&	0.852 	&	0.729 	&	0.722 	\\	
CNN-J(+)	&	0.851 	&	0.851 	&	\textbf{0.738} 	&	\textbf{0.732} 	\\	\hline
FM	&	0.765 	&	0.764 	&	0.607 	&	0.540 	\\	
CFM	&	0.826 	&	0.822 	&	0.785 	&	0.784 	\\	
PFM	&	\textbf{0.850}	&	\textbf{0.850}	&	\textbf{0.789}	&	\textbf{0.788}	\\
\hline
\end{tabular}
\end{center}
\caption{\label{tab:movie} Document and snippet classification results for movie. The numbers in bold highlight the best results in FMs and other baselines.}
\end{table} 

%
%
\renewcommand{\tabcolsep}{1.5pt} 

\begin{table}
\small
\begin{center}
\begin{tabular}{|c|c|c|c|c|c|c|c|c|}

\hline
\multicolumn{1}{|c|}{\rule{0pt}{8pt} \textbf{}} & \multicolumn{2}{|c|}{\rule{0pt}{8pt} \textbf{Document}} & \multicolumn{2}{|c|}{\rule{0pt}{8pt} \textbf{Snippet}} \\
\hline
\rule{0pt}{8pt} \textbf{Models}  & \textbf{Accuracy}  & \textbf{F1-Score}  & \textbf{Accuracy}  & \textbf{F1-Score}
\\
\hline

LR-BoW	&	0.838 	&	0.837 	&	0.823 	&	0.823 	\\	
SVM-BoW 	&	0.837 	&	0.836 	&	\textbf{0.824} 	&	\textbf{0.824} 	\\	
SVM-Poly & 0.823 & 0.823 & 0.740 & 0.733 \\ 
LR-WE	&	0.767 	&	0.767 	&	0.737 	&	0.733 	\\	
SVM-WE	&	0.767 	&	0.766 	&	0.736 	&	0.732 	\\	
CNN-S	&	0.754 	&	0.755 	&	0.731 	&	0.731 	\\	
CNN-J	&	0.823 	&	0.823 	&	0.799 	&	0.799 	\\	
CNN-S(+)	&	\textbf{0.859} 	&	\textbf{0.859} 	&	0.815 	&	0.815 	\\	
CNN-J(+)	&	0.849 	&	0.849 	&	0.810 	&	0.810 	\\	\hline
FM	&	0.786 	&	0.786 	&	0.745 	&	0.731 	\\	
CFM	&	0.835 	&	0.833 	&	0.822 	&	0.821 	\\	
PFM	&	\textbf{0.842}	&	\textbf{0.841}	&	\textbf{0.833}	&	\textbf{0.833}	\\	
\hline
\end{tabular}
\end{center}
\caption{\label{tab:cellphone} Document and snippet classification results for cellphone. The numbers in bold highlight the best results in FMs and other baselines.}
\end{table} 

%
%
\renewcommand{\tabcolsep}{1.5pt} 

\begin{table}
\small
\begin{center}
\begin{tabular}{|c|c|c|c|c|c|c|c|c|}

\hline
\multicolumn{1}{|c|}{\rule{0pt}{8pt} \textbf{}} & \multicolumn{2}{|c|}{\rule{0pt}{8pt} \textbf{Document}} & \multicolumn{2}{|c|}{\rule{0pt}{8pt} \textbf{Snippet}} \\
\hline
\rule{0pt}{8pt} \textbf{Models}  & \textbf{Accuracy}  & \textbf{F1-Score}  & \textbf{Accuracy}  & \textbf{F1-Score}
\\
\hline

LR-BoW	&	0.809 	&	0.808 	&	0.812 	&	0.812 	\\	
SVM-BoW	&	0.808 	&	0.807 	&	0.811 	&	0.810 	\\	
SVM-Poly & 0.778 & 0.777 & 0.594 & 0.520 \\ 
LR-WE	&	0.743 	&	0.742 	&	0.690 	&	0.689 	\\	
SVM-WE	&	0.742 	&	0.741 	&	0.683 	&	0.681 	\\	
CNN-S	&	0.787 	&	0.787 	&	0.739 	&	0.739 	\\	
CNN-J	&	0.823 	&	0.822 	&	0.785 	&	0.785 	\\	
CNN-S(+)	&	0.827 	&	0.827 	&	0.805 	&	0.804 	\\	
CNN-J(+)	&	\textbf{0.846} 	&	\textbf{0.846} 	&	\textbf{0.818} 	&	\textbf{0.818} 	\\	\hline
FM	&	0.790 	&	0.788 	&	0.698 	&	0.674 	\\	
CFM	&	\textbf{0.839}	&	\textbf{0.839}	&	0.838 	&	0.837 	\\	
PFM	&	\textbf{0.839} 	&	\textbf{0.839} 	&	\textbf{0.8424}	&	\textbf{0.842}	\\	
\hline
\end{tabular}
\end{center}
\caption{\label{tab:restaurant} Document and snippet classification results for restaurant. The numbers in bold highlight the best results in FMs and other baselines.}
\end{table} 


\subsection{Experimental Results and Analysis}
The experimental results are given in Table~\ref{tab:movie}, ~\ref{tab:cellphone} and ~\ref{tab:restaurant}. In every table, the left hand side shows the accuracy and F1-Score for document-level sentiment classification and the right hand side shows the snippet-level ones. A table consists of two parts, where the lower part belongs to FM models (FMs) and the upper part presents the baselines. The highest scores are marked in bold for both parts.


First, we have the following observations at the document level (from first two columns in tables):

\begin{enumerate}[topsep=0pt,leftmargin=*]
\item Most models have competitive performance (except LR-WE, SVM-WE, CNN-S and CNN-J), which implies that when rich information is available in a full review, simply summarizing its overall sentiment orientation can alleviate the problem of lacking SO word interaction. 

\item Our proposed models CFM and PFM are comparable to state-of-the-art baselines. While a small vector length and simplified parameters are used in CFM and PFM for learning, their performance is close to CNN-S/J(+), which is encouraging.

\item CFM and PFM outperform FM in all data sets, which shows their superiority and rationality in sentiment analysis.

\end{enumerate}

Second, we have the following observations at the snippet level (from last two columns in tables):

\begin{enumerate}[topsep=0pt,leftmargin=*]
\item Our proposed models CFM and PFM dramatically outperform other baselines significantly in this fine-grained setting. In addition, they can consistently achieve good results on three data sets.

\item Compared to CNN-S/J(+), CFM and PFM have better performance, even when these CNNs use bigger size vectors (length 300). It is worth noting that these two CNNs actually achieve very good results at the document level (see the left two columns) and their parameters are well fit; however, they do not perform well at the snippet level as ours, which indicates the effectiveness of capturing SO word interaction.


\item FM preforms very poorly and we can see the significant improvement gain from CFM and PFM which demonstrates their effectiveness for fine-grained sentiment analysis. 
\end{enumerate}

Third, we have the following further observations:
\begin{enumerate}[topsep=0pt,leftmargin=*]
\item Considering the performances on both two settings, we can see the robustness of PFM and also CFM, where they can achieve consistently stable results.

\item We also tried SVM with different kernels including SVM-Poly, but the linear SVM achieves the best results consistently. It has also been reported by researchers that the linear kernel performs the best for binary text classification.~\cite{joachims1998text, colas2006comparison,fei2015social}. \\

\end{enumerate}



\section{Related Work}
\label{sec:relatedwork}
In machine learning context, Bag-of-Words models were first used for building classifiers~\cite{pang2002thumbs} for sentiment classification.
Later, dense low-dimensional word vector becomes a better alternative~\cite{,blei2003latent} for word representation. Recently word embeddings like skip-gram model~\cite{mikolov2013efficient} have shown their superiority in many NLP tasks~\cite{turian2010word}.~\cite{maas2011learning} first introduced a topic model variant to jointly encode sentiment and semantic information; later with the development of CNN~\cite{collobert2011natural} in text mining, joint CNN models~\cite{kim2014convolutional,tang2014learning} achieve better and state-of-the-art results. But they did not conduct fine-grained analysis at the snippet level.~\cite{tang2016aspect,li2017deep} performed aspect-level sentiment classification using aspect labels for training and testing, which is essentially different from our task. None of the above work explicitly captures SO word interaction.   

Another related work is from~\citet{he2016pairwise} who considered the word interaction problem but aimed at mapping similar word interactions across different sentences.~\citet{johnson2014effective} inspected position information but it is not for modeling word interactions. 

The concept of SO word interaction is related to sentiment negation/shifter theory~\cite{polanyi2006contextual}, contextual polarity~\cite{wilson2009recognizing} and sentiment composition related works~\cite{choi2008learning,moilanen2007sentiment,neviarouskaya2009compositionality}. However, they do not target at learning a joint model with information encoded in SWI vectors like us. Also, we do not use external resources like NLP parser~\cite{socher2013recursive,naseem2010using} to help infer sentiment information. 

Factorization Machine (~\cite{rendle2010factorization} is a general approach that learns feature conjunctions. It is widely used for recommender system~\cite{jahrer2012ensemble,petroni2015core,juan2016field} but existing FMs are not suitable for sentiment analysis. we also compared the classical FM in our experiments.

\section{Conclusion}
This paper introduced a framework that can explicitly capture sentiment-oriented word interaction by learning a set of sentiment-sensitive Word Interaction (SWI) vectors. Specifically, two new models were developed, namely, Contextual Factorization Machine (CFM) and Position-aware Factorization Machine (PFM). They benefit fine-grained analysis at the snippet level and also simplify the parameter learning. Extensive experimental results show their effectiveness.

\bibliography{cpfm_sentiment_fm}
\bibliographystyle{acl_natbib}

\end{document}